\pdfoutput=1

\documentclass{article}
\usepackage{latexsym}

\usepackage{spconf,amsmath,graphicx}
\usepackage{times}
\usepackage{amsmath}
\usepackage{algorithm}
\usepackage{algpseudocode}
\usepackage{diagbox}
\usepackage{latexsym}
\usepackage{enumitem}
\usepackage{siunitx}
\usepackage{helvet}  
\usepackage{courier}  
\usepackage{url}  
\usepackage{graphicx}  
\usepackage{courier}
\usepackage{amsmath}
\usepackage[normalem]{ulem}
\useunder{\uline}{\ul}{}
\usepackage{graphicx}
\usepackage{multirow}
\usepackage{mathbbol}
\usepackage{caption}
\usepackage{CJKutf8}
\usepackage{verbatim}
\usepackage{float}
\usepackage{booktabs}
\usepackage{array}
\usepackage{times}
\usepackage{csquotes}
\usepackage{latexsym}
\usepackage{tablefootnote}
\usepackage[normalem]{ulem}
\usepackage{todonotes}
\usepackage{subcaption}
\usepackage{tikz}
\usepackage{xcolor}
\usepackage{hhline}
\usepackage{placeins}

\title{Parameter Efficient Audio Captioning with faithful guidance using Audio-Text Shared Latent Representation}
\name{Arvind Krishna Sridhar, Yinyi Guo, Erik Visser, Rehana Mahfuz}
\address{Qualcomm Technologies}
\begin{document}
\setlength{\abovedisplayskip}{0pt}
\setlength{\belowdisplayskip}{0pt}
\setlength{\abovedisplayshortskip}{0pt}
\setlength{\belowdisplayshortskip}{0pt}

%
\maketitle
\begin{abstract}
There has been significant research on developing pretrained transformer architectures for multimodal-to-text generation tasks. Albeit performance improvements, such models are frequently overparameterized, hence suffer from hallucination and large memory footprint making them challenging to deploy on edge devices. In this paper, we address both these issues for the application of automated audio captioning. First, we propose a data augmentation technique for generating hallucinated audio captions and show that similarity based on an audio-text shared latent space is suitable for detecting hallucination. Then, we propose a parameter efficient inference time faithful decoding algorithm that 
enables smaller audio captioning models with performance equivalent to larger models trained with more data. During the beam decoding step, the smaller model
utilizes an audio-text shared latent representation to semantically align the generated text with corresponding input audio. Faithful guidance is introduced into the beam probability by incorporating the cosine similarity between latent representation projections of greedy rolled out intermediate beams and audio clip. We show the efficacy of our algorithm on benchmark datasets and evaluate the proposed scheme against baselines using conventional audio captioning and semantic similarity metrics while illustrating tradeoffs between performance and complexity.

\end{abstract}
\section{Introduction}
\vspace{-1mm}
In recent years, there has been extensive research on pushing the boundaries of multimodal-to-text generation tasks like image captioning \cite{rohrbach-etal-2018-object}, audio captioning (\cite{Ye2021ImprovingTP},  \cite{chen22h_interspeech}) etc. Although there has been significant research in improving model performance, two major bottlenecks, hallucination \cite{10.1145/3571730} and large memory footprint\cite{ganesh2021BERTissues}, remain that inhibit the wide scale adoption of such tasks on constrained computing devices. In \cite{10.1145/3571730}, hallucination is defined as ”the generated content that is nonsensical or unfaithful to the provided source content”. Their survey documents research on hallucination in multimodal to text generation tasks such as abstractive summarization and vision-language generation.
Second, improved performance of large pretrained transformer models on evaluation benchmarks comes at the cost of larger memory footprint\cite{ganesh2021BERTissues}. These models are often over-parameterized with respect to the task they are solving, thus necessitating architectural innovations to tackle computational complexity during deployment. \\
In this paper, we address both of these bottlenecks by proposing automated audio captioning models retrofitted with a hallucination detection and mitigation at decoding stage - the task of generating a relevant audio caption given an audio. To the best of our knowledge, we are the first to make the following contributions in this domain. First, we propose a data augmentation technique to generate hallucinated audio captions using existing audio captioning datasets by leveraging large language models. Second, we provide an intuitive reasoning on why existing audio captioning metrics are not suitable for detecting hallucination. Instead we argue that acoustic similarity of captions needs to be taken into account and introduce a hallucination metric based on an audio-text shared latent representation. Third, we propose an inference time hallucination mitigation algorithm where similarity of the intermediate beams with respect to the input audio is used to faithfully guide the beams during beam decoding. We show that our retrofitting method enables smaller audio captioning models with performance equivalent to much larger models.
\vspace{-4mm}
\section{Related Works}
\vspace{-3mm}
 \subsection{Audio Captioning}
\vspace{-3mm}
Conventional audio captioning systems are based on encoder-decoder architectures where the encoder captures the temporal and acoustic information of the audio and the decoder generates the caption auto-regressively(\cite{chen22h_interspeech}, \cite{mei2023WavCaps}). In addition to the encoded audio representations, keywords are extracted from the audio and provided to the decoder to achieve grounded guidance(\cite{Ye2021ImprovingTP}, \cite{chen22h_interspeech}).
 The study on lexical diversity and similarity of captions\cite{MartinMorato2021} shows that different annotators interpret the same audio using different vocabulary. In this work, we analyze  the properties of SOTA audio captioning evaluation metrics when used for semantic and acoustic similarity detection and find that similarity metrics based on audio-text shared latent representation are better suited for such tasks.


  \begin{table*}[!ht]
    \centering
    \begin{tabular}{|p{4.9cm}|p{4.4cm}|p{7.5cm}|}
    \hline
        Original Caption & Injecting Audio Tags & Hallucinated Caption\\ \hline
        A campfire in the night time with crickets and other bugs making noise in the background. & Bird, Speech, Outside, urban or manmade & A nighttime campfire with crickets and other bugs chirping in the background, accompanied by the sound of human speech. \\ \hline
        A crowd of people and a child begin talking as cars beep in the background and then the crowd cheers. & Outside, urban or manmade, Singing, Insect & A child is playing outside in an urban area while singing and insects are heard in the background. \\ \hline
    \end{tabular}
    \caption{Hallucinated audio captions generated using the proposed data augmentation technique.}
    \label{Table:5}
\end{table*}
\begin{table*}[!ht]
    \centering
    \begin{tabular}{|l|l|l|l|l|l|l|l|l|l|}
    \hline
        Data$\backslash$Metrics &  BLEU 1 & METEOR & ROUGE L & CIDER & SPICE & SentBert & $\textrm{CLAPScore}_{tt}$ \\ \hline
        Hallucinated  & 0.4338 & 0.223 & 0.3773 & 0.2077 & 0.1491 & 0.5414 & 0.3609 \\ \hline
        Non Hallucinated & 0.5798 & 0.3408 & 0.5663 & 0.8031 & 0.217 & 0.8627 & 0.701 \\ \hline
    \end{tabular}
    \caption{Performance of evaluation metrics on audio captioning hallucination benchmark.}
    \label{Table:1}
    \vspace*{-5mm}
\end{table*}

 \subsection{Hallucination in natural language generation and its detection metrics}
 Mitigating hallucination via training\cite{balachandran-etal-2022-correcting} and during inference time(\cite{sridhar2022improved}, \cite{king-etal-2022-dont}) is an ongoing focus of natural language research. FactEdit\cite{balachandran-etal-2022-correcting} performs rewriting of the generated summary to avoid hallucination while \cite{sridhar2022improved}, \cite{king-etal-2022-dont}  propose decoding techniques to reduce hallucination on the fly during decoding time. In \cite{rohrbach-etal-2018-object}, hallucination is investigated in image captioning and a CHAIR metric is proposed that computes the ratio of generated objects to objects found in image and ground-truth captions. Further, it is observed in \cite{dai-etal-2023-plausible} that models that perform well on standard captioning metrics like CIDER\cite{mei2023WavCaps} still produce unfaithful texts. In \cite{10.1007/978-3-031-26316-3_37}, the problem of hallucination in video captioning is studied. To our knowledge, we are the first to study hallucination in the audio captioning domain.
\section{Methodology}
    We divide our proposed methodology into three sections. First, we explain our novel data augmentation technique to generate hallucinated audio captions. Second, we investigate and propose a hallucination metric to measure hallucinations in audio captioning. Third, we explain in detail our proposed faithful decoding algorithm.
\vspace{-3mm}
\subsection{Generating hallucinated data}
\label{sec:GenHalldata}
In order to investigate and study hallucination in audio captioning, we introduce a data augmentation technique that removes zero or more audio events from the original caption and gradually augments it with similar/dissimilar audio events. Table \ref{Table:5} illustrates generated hallucinated captions using this method. First, we randomly select 50 examples from the Clotho\cite{clotho} dataset. We randomly select one of the five ground truth captions as original caption and paraphrase it using vicuna checkpoint\cite{zheng2023judging} of LLaMA 2\cite{touvron2023llama} to generate the non hallucinated data points.  Second, we retrieve audio tags using Audio spectrogram transformer\cite{gong21b_interspeech} for the corresponding audio clip. We assume the audio tags with low classification scores are acoustically dissimilar from the input audio. So, we randomly select three tags in the range of 30-40(dissimilar) from a list of descending order ranked audio tags based on classification score. We generate a modified audio caption using LLaMA 2\cite{touvron2023llama} with in-context learning by providing the original audio caption, the audio tags to inject and a few examples.

\begin{figure*}[t!]
\begin{minipage}[b]{1.0\linewidth}
  \centering
  \centerline{
  \includegraphics[width=1.0\textwidth,scale=0.5]{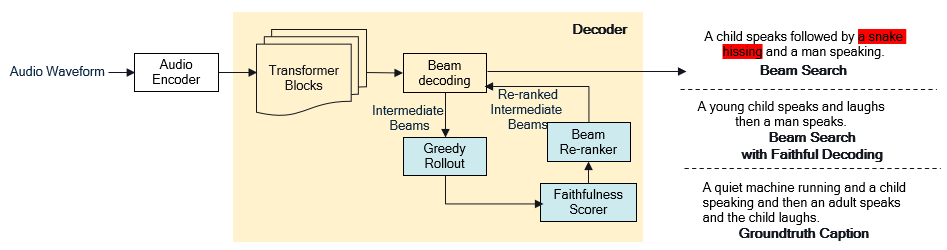}
  }
\end{minipage}
\caption{Proposed faithful decoding for audio captioning. The proposed components are colored light sky blue.}
\label{fig:res}
\end{figure*}

\vspace{-3mm}
\subsection{CLAPScore as hallucination metric}
\label{sec:hallucination metric}
For a metric to be considered suitable for hallucination detection, it should have two properties - 1) detect false positive audio events in the caption 2) account for acoustic similarity between audio events. Contrastive Language Audio Pretraining (CLAP) \cite{laionclap2023} uses an audio and text encoder to learn a shared embedding space using contrastive learning which shows SOTA results for downstream tasks like audio retrieval, audio captioning \cite{mei2023WavCaps}. We introduce CLAPScore to compute the cosine similarity over audio-text and text-text pairs as shown in Equation 1.

\begin{table*}[!ht]
    \centering
    \begin{tabular}{|p{7.1cm}|p{0.9cm}|p{1.3cm}|p{1.2cm}|p{0.9cm}|p{0.8cm}|p{1.1cm}||p{1cm}|}
    \hline
        Example pair$\backslash$Metrics & BLEU-1 & METEOR & ROUGE-L& CIDER & SPICE & SentBert  & \begin{tabular}[c]{@{}c@{}}$\textrm{CLAP}$\\$\textrm{Score}_{tt}$\end{tabular} \\ \hline
       Horse is trotting., Someone is tapping on a surface. & 0.3033 & 0.0702 & 0.193 & 0 & 0 & -0.0572 & 0.6476 \\ \hline
        Horse is trotting., Someone is running on wood. & 0.1947 & 0.0379 & 0.2179 & 0 & 0 & 0.2262 & 0.7223 \\ \hline \hline
        Crowd is applauding the performer., Crows is silent after the performer's show. & 0.439 & 0.2374 & 0.5908 & 0.0 & 0.4444 & 0.3596 & 0.3836 \\ \hline
    \end{tabular}
    \caption{Qualitative analysis of evaluation metrics on acoustic similar and dissimilar audio captions. The first two pairs are similar acoustic captions while the third pair is an example of dissimilar acoustic caption.}
    \label{Table:2}
    \vspace*{-5mm}
\end{table*}

\begin{align*}
\small
\vspace*{-4cm}
\textrm{CLAPScore} = \frac{x_{A}\cdot x_{B}}{\vert x_{A}\vert \cdot \vert x_{B}\vert} \tag{1}
\end{align*} 
\vspace{1mm}
Where $x_{A}$ and $x_{B}$ represents CLAP projected representations of audio or text. Hereon, we denote the  $\textrm{CLAPScore}$ between audio and text as $\textrm{CLAPScore}_{at}$ while between two texts as $\textrm{CLAPScore}_{tt}$.\\
We adopt the audio captioning metrics used by \cite{mei2023WavCaps} including CIDER, SPICE, BLEU, METEOR, ROUGE L scores and text semantic similarity based metric SentBert\cite{reimers-gurevych-2019-sentence}. For the first condition, we show the performance of standard metrics on the generated hallucinated and non-hallucinated, as described in section \ref{sec:GenHalldata}, serving as benchmark. For the second condition, acoustic similarity detection needs to be an inherent property of the metric. Since it is difficult to categorize audio clips into acoustically similar pairs systematically, we perform a qualitative study as shown in Table \ref{Table:2}. From Table \ref{Table:1} and \ref{Table:2}, We can observe that none of the metrics except $\textrm{CLAPScore}_{tt}$ satisfy both these properties. Although the standard metrics perform well on detecting text hallucinations in Table \ref{Table:1}, they are unable to distinguish between acoustic similar and dissimilar audio captions as shown in Table \ref{Table:2}. This is due to them considering only text embedding space during computation. The audio captions - "Horse is trotting." and "Someone is walking on the wood." might seem very different in the text domain whereas in acoustic domain both sounds are acoustically similar to hear. Hence, such audio events should be less penalized by a hallucination metric compared to the acoustically contrasting audio caption pairs such as (”Crowd is applauding the performer.”, ”Crowd is silent after the performer’s show”).   Since $\textrm{CLAPScore}_{tt}$ satisfies both the properties, in this paper, we adopt CLAP as the hallucination metric.
\vspace{-3mm}
\subsection{Faithful decoding algorithm}
In this section, we propose a faithful decoding algorithm that utilizes the ability of CLAP to detect hallucinations (shown in Section \ref{sec:hallucination metric}) during inference time.
\vspace{-3mm}
\subsubsection{Greedy rollout}
\vspace{-2mm}
Beam search performs a breadth first search at each decoding step with limited branches from Begin of sentence (BOS) to End of sentence (EOS) \cite{beamsearch}. Each path from BOS to EOS are called hypothesis. During beam decoding process, only partial hypothesis or intermediate beams (paths that start at BOS and end before EOS) are available for re-ranking. To compare the intermediate beams against the input audio, we complete them using greedy search\cite{beamsearch}. Greedy search samples the token with highest probability at every decoding step as shown in Equation 2. This serves us as a look ahead for how the beam would pan out in case it goes down that direction.

\begin{align*}
\small
\vspace*{-4cm}
P_\Theta (y|x) = \prod_{t=1}^{|y|}P_\Theta (y_t | x, y_{<t}),
 \tag{2}
 \label{eq:greedysearch}
\end{align*} 
where $x$ is the input and $y_t$ is the word generated at $t^{th}$ decoding step.
\vspace{-3mm}
\subsubsection{Faithfulness scorer}
Next, we compute the relevance of greedy rolled out beam with the input audio by taking the CLAP projections of beam text and audio. We normalize the projections and take cosine similarity to compute $\textrm{CLAPScore}_{at}$ as the distance between greedy rolled out beam and audio in the shared embedding space (Equation 1).
\begin{table*}[!ht]
    \centering
    \begin{tabular}{|p{3cm}|l|l|l|l|p{1.2cm}|p{1.2cm}|p{1.8cm}|}
    \hline
        Dataset+Sampling & BLEU 1 & METEOR & ROUGE L & CIDER & SPICE & SentBert  & $\textrm{CLAPScore}_{tt}$\\ \hline
        AC+Beam & 0.6515 & 0.2174 & \textbf{0.4597} & 0.6101 & 0.1597 & 0.7035 & 0.8208 \\ \hline
        AC+Clap Beam & 0.6323 & 0.2267 & 0.4455 & 0.6249 & 0.1595 & 0.7298 & 0.8242 \\ \hline
        AC+Htsatbert Beam & \textbf{0.6605} & \textbf{0.2330} & 0.4584 & \textbf{0.6593} & \textbf{0.1677} & \textbf{0.7554} & \textbf{0.8618} \\ \hline \hline
        C+Beam & 0.5496 & 0.1706 & \textbf{0.3738} & 0.3745 & 0.1141 & 0.6098 & 0.7404 \\ \hline                
        C+Clap Beam & 0.4931 & 0.1711 & 0.3412 & 0.3184 & 0.1185 & 0.6405 & 0.7814\\ \hline
        C+Htsatbert Beam & \textbf{0.5570} & \textbf{0.1793} & 0.3695 & \textbf{0.4187} & \textbf{0.1283} & \textbf{0.6641} & \textbf{0.7995} \\ \hline
    \end{tabular}
    \caption{Main Result: Performance of baseline with proposed faithful decoding on AudioCaps(AC) and Clotho(C) datasets. The bold numbers signify best performance on corresponding dataset. }
    \vspace*{-5mm}
    \label{Table:Mainresult}
\end{table*}
\vspace*{-3mm}
\subsubsection{Beam re-ranker}
To incorporate the $\textrm{CLAP}_{score} $ into beam decoding we weight it over the model probability $\textrm{P}_i$ to compute $\textrm{P}_{weighted}$ (Equation 3). The modified probability $\textrm{P}_{weighted}$ ensures to guide the beam to explore regions faithful to input audio thereby reducing hallucination.
\begin{align*}
\small
\vspace*{4mm}
\textrm{P}_{weighted} = (1-\alpha) \textrm{P}_i  + \alpha \textrm{CLAPScore}_{at} \tag{3}
\end{align*}
where $P_{i}$ denotes the model probability for $i^{ith}$ token.

\begin{table}[!ht]
    \centering
    \begin{tabular}{|p{2.9cm}|p{0.9cm}|p{0.8cm}|p{1.1cm}|p{1cm}|}
    \hline
        Dataset+Sampling & CIDER & SPICE & SentBert & 
        \begin{tabular}[c]{@{}c@{}}$\textrm{CLAP}$\\$\textrm{Score}_{tt}$\end{tabular}\\ \hline
        AC+Beam & \textbf{0.7904} & 0.1821 & \textbf{0.7926} & 0.8849\\ \hline
        AC+Clap Beam & 0.7252 & 0.1806 & 0.7847 & 0.8727 \\ \hline
        AC+Htsatbert Beam& 0.7232 & \textbf{0.1828} & 0.7915 & \textbf{0.8859}\\ \hline \hline
        C+Beam & \textbf{0.4794} & 0.1323 & \textbf{0.6766} & \textbf{0.8175}\\ \hline 
        C+Clap Beam & 0.4394 & 0.1317 & 0.6735 & 0.8079 \\ \hline
        C+Htsatbert Beam & 0.4462 & \textbf{0.1324} & 0.6744 & 0.8154\\ \hline
        
    \end{tabular}
    \caption{Performance of HTSAT-BART\cite{mei2023WavCaps} with proposed faithful decoding on AudioCaps(AC) and Clotho(C) datasets. The bold numbers signify best performance on corresponding dataset.}
    \label{Table:htsatbartresult}
    \vspace{-6mm}
\end{table}

\vspace{-2mm}
\section{Experiments}
\vspace{-2mm}
\subsection{Datasets}
We trained the models on Clotho\cite{clotho} and AudioCaps\cite{kim-etal-2019-audiocaps} audio captioning datasets. Clotho audio captioning dataset comprises of 4981 audio samples with each sample accompanied by 5 human written captions. The duration of the audio clips are in the range of 15 to 30 sec. Audiocaps\cite{kim-etal-2019-audiocaps} consist of 46k human written captions obtained via crowdsourcing with 10 sec duration for each audio clip. For both the datasets, we use the test set for our evaluation. For fixed length transformer encoders like HTSAT BERT\cite{mei2023WavCaps}, we truncate the audio sample to 10 sec and resample at the rate of 32000.
\vspace{-3mm}
\subsection{Experiment Setup}
We demonstrate the performance of our model against  \cite{Mei2021AnEB} chosen especially due to its small size. The model is trained on Clotho and AudioCaps datasets from scratch and evaluated correspondingly. It consists of CNN10 PANN pretrained as the encoder and a stack of two transformer decoder layers as decoder. To compare against large models, we show our audio captioning performance on HTSAT-BART\cite{mei2023WavCaps} which consists of HTSAT as audio encoder and BART\cite{mei2023WavCaps} as decoder. For the shared embedding space to get projections, we use CLAP\cite{laionclap2023} and HTSAT BERT\cite{mei2023WavCaps}. We use the LAION checkpoint\cite{laionclap2023} for CLAP and WavCaps checkpoint for HTSAT BERT\cite{mei2023WavCaps}. We use 0.8 and 0.6 as $\alpha$ value for experiments in Table \ref{Table:Mainresult} and Table \ref{Table:htsatbartresult}. Clap Beam and Htsatbert Beam refers to the proposed faithful decoding algorithm with CLAP and Htsatbert as the shared embedding space to project audio and text respectively.
\vspace{-3mm}
\section{Results}
\vspace{-1mm}
From Table \ref{Table:Mainresult}, the improvements on $\textrm{CLAPScore}_{tt}$ by 0.04 and 0.06 for AudioCaps and Clotho datasets indicate reduced hallucinations. We also observe that the proposed faithful decoding improves the performance of baseline model across all metrics except ROUGE L.  This demonstrates that the proposed faithful decoding not only reduces hallucination but improves overall caption quality for smaller models. As a sanity check and to compare the performance against larger models, we perform the same experiments on HTSAT-BART\cite{mei2023WavCaps}. In Table \ref{Table:htsatbartresult}, the proposed faithful decoding slightly outperforms on SPICE and $\textrm{CLAPScore}_{tt}$ for AudioCaps while not causing a significant overall change in evaluation metrics. This is expected since the HTSAT-BART, being a large model and trained on Wavcaps\cite{mei2023WavCaps} a much larger dataset (630k samples) than Clotho and AudioCaps, does not get extra useful information from the shared embedding space.

\vspace{-3mm}
\section{Conclusion}
\vspace{-1mm}
We investigated the hallucination problem in audio captioning and proposed a new hallucination augmentation technique which will aid in future research of hallucination mitigation algorithms and metrics. Then, we showed that cosine similarity on audio-text shared embedding is a good hallucination metric. With no further finetuning, we proposed an inference time faithful decoding algorithm that utilizes shared embedding space to guide the beams during decoding time. In the future, we plan to develop a hallucination loss for finetuning stage.
\vspace{-3mm}





\bibliographystyle{IEEEbib}
\bibliography{refs}

\begin{thebibliography}{10}

\bibitem{rohrbach-etal-2018-object}
Anna Rohrbach, Lisa~Anne Hendricks, Kaylee Burns, Trevor Darrell, and Kate
  Saenko,
\newblock ``Object hallucination in image captioning,''
\newblock in {\em Proceedings of the 2018 Conference on Empirical Methods in
  Natural Language Processing}. Oct.-Nov., pp. 4035--4045, ACL.

\bibitem{Ye2021ImprovingTP}
Zhongjie Ye, Helin Wang, Dongchao Yang, and Yuexian Zou,
\newblock ``Improving the performance of automated audio captioning via
  integrating the acoustic and semantic information,''
\newblock in {\em Workshop on Detection and Classification of Acoustic Scenes
  and Events}, 2021.

\bibitem{chen22h_interspeech}
Kun Chen, Jun Wang, Feng Deng, and Xiaorui Wang,
\newblock ``{iCNN-Transformer: An improved CNN-Transformer with Channel-spatial
  Attention and Keyword Prediction for Automated Audio Captioning},''
\newblock in {\em Proc. Interspeech 2022}, 2022, pp. 4167--4171.

\bibitem{10.1145/3571730}
Ziwei Ji, Nayeon Lee, Rita Frieske, Tiezheng Yu, Dan Su, Yan Xu, Etsuko Ishii,
  Ye~Jin Bang, Andrea Madotto, and Pascale Fung,
\newblock ``Survey of hallucination in natural language generation,''
\newblock {\em ACM Comput. Surv.}, vol. 55, no. 12, 2023.

\bibitem{ganesh2021BERTissues}
Prakhar Ganesh, Yao Chen, Xin Lou, Mohammad~Ali Khan, Yin Yang, Hassan Sajjad,
  Preslav Nakov, Deming Chen, and Marianne Winslett,
\newblock ``{Compressing Large-Scale Transformer-Based Models: A Case Study on
  BERT},''
\newblock {\em Transactions of the Association for Computational Linguistics},
  vol. 9, pp. 1061--1080, 2021.

\bibitem{mei2023WavCaps}
Xinhao Mei, Chutong Meng, Haohe Liu, Qiuqiang Kong, Tom Ko, Chengqi Zhao,
  Mark~D. Plumbley, Yuexian Zou, and Wenwu Wang,
\newblock ``Wav{C}aps: A {ChatGPT}-assisted weakly-labelled audio captioning
  dataset for audio-language multimodal research,''
\newblock {\em arXiv preprint arXiv:2303.17395}, 2023.

\bibitem{MartinMorato2021}
Irene {Martin Morato} and Annamaria Mesaros,
\newblock ``Diversity and bias in audio captioning datasets,''
\newblock in {\em Proceedings of the 6th Workshop on Detection and Classication
  of Acoustic Scenes and Events (DCASE 2021)}, pp. 90--94.

\bibitem{balachandran-etal-2022-correcting}
Vidhisha Balachandran, Hannaneh Hajishirzi, William Cohen, and Yulia Tsvetkov,
\newblock ``Correcting diverse factual errors in abstractive summarization via
  post-editing and language model infilling,''
\newblock in {\em Proceedings of the 2022 Conference on Empirical Methods in
  Natural Language Processing}. pp. 9818--9830, ACL.

\bibitem{sridhar2022improved}
Arvind~Krishna Sridhar and Erik Visser,
\newblock ``Improved beam search for hallucination mitigation in abstractive
  summarization,''
\newblock {\em arXiv preprint arXiv:2212.02712}, 2022.

\bibitem{king-etal-2022-dont}
Daniel King, Zejiang Shen, Nishant Subramani, Daniel~S. Weld, Iz~Beltagy, and
  Doug Downey,
\newblock ``Don{'}t say what you don{'}t know: Improving the consistency of
  abstractive summarization by constraining beam search,''
\newblock in {\em Proceedings of the 2nd Workshop on Natural Language
  Generation, Evaluation, and Metrics (GEM)}. 2022, pp. 555--571, ACL.

\bibitem{dai-etal-2023-plausible}
Wenliang Dai, Zihan Liu, Ziwei Ji, Dan Su, and Pascale Fung,
\newblock ``Plausible may not be faithful: Probing object hallucination in
  vision-language pre-training,''
\newblock in {\em Proceedings of the 17th Conference of the European Chapter of
  the Association for Computational Linguistics}. 2023, pp. 2136--2148, ACL.

\bibitem{10.1007/978-3-031-26316-3_37}
Nasib Ullah and Partha~Pratim Mohanta,
\newblock ``Thinking hallucination for video captioning,''
\newblock in {\em Computer Vision – ACCV 2022: 16th Asian Conference on
  Computer Vision, 2022, Part IV}. p. 623–640, Springer-Verlag.

\bibitem{clotho}
Konstantinos Drossos, Samuel Lipping, and Tuomas Virtanen,
\newblock ``Clotho: an audio captioning dataset,''
\newblock in {\em ICASSP 2020 - 2020 IEEE International Conference on
  Acoustics, Speech and Signal Processing (ICASSP)}, 2020, pp. 736--740.

\bibitem{zheng2023judging}
Lianmin Zheng, Wei-Lin Chiang, Ying Sheng, Siyuan Zhuang, Zhanghao Wu, Yonghao
  Zhuang, Zi~Lin, Zhuohan Li, Dacheng Li, Eric.~P Xing, Hao Zhang, Joseph~E.
  Gonzalez, and Ion Stoica,
\newblock ``Judging llm-as-a-judge with mt-bench and chatbot arena,'' 2023.

\bibitem{touvron2023llama}
Hugo Touvron, Louis Martin, Kevin Stone, Peter Albert, Amjad Almahairi, Yasmine
  Babaei, Nikolay Bashlykov, Soumya Batra, Prajjwal Bhargava, Shruti Bhosale,
  et~al.,
\newblock ``Llama 2: Open foundation and fine-tuned chat models,''
\newblock {\em arXiv preprint arXiv:2307.09288}, 2023.

\bibitem{gong21b_interspeech}
Yuan Gong, Yu-An Chung, and James Glass,
\newblock ``{AST: Audio Spectrogram Transformer},''
\newblock in {\em Proc. Interspeech 2021}, 2021, pp. 571--575.

\bibitem{laionclap2023}
Yusong Wu*, Ke~Chen*, Tianyu Zhang*, Yuchen Hui*, Taylor Berg-Kirkpatrick, and
  Shlomo Dubnov,
\newblock ``Large-scale contrastive language-audio pretraining with feature
  fusion and keyword-to-caption augmentation,''
\newblock in {\em IEEE International Conference on Acoustics, Speech and Signal
  Processing, ICASSP}, 2023.

\bibitem{reimers-gurevych-2019-sentence}
Nils Reimers and Iryna Gurevych,
\newblock ``Sentence-{BERT}: Sentence embeddings using {S}iamese
  {BERT}-networks,''
\newblock in {\em Proceedings of the 2019 Conference on Empirical Methods in
  Natural Language Processing and the 9th International Joint Conference on
  Natural Language Processing (EMNLP-IJCNLP)}. pp. 3982--3992, ACL.

\bibitem{beamsearch}
Clara Meister, Tim Vieira, and Ryan Cotterell,
\newblock ``{Best-First Beam Search},''
\newblock {\em Transactions of the Association for Computational Linguistics},
  vol. 8, pp. 795--809, 2020.

\bibitem{kim-etal-2019-audiocaps}
Chris~Dongjoo Kim, Byeongchang Kim, Hyunmin Lee, and Gunhee Kim,
\newblock ``{A}udio{C}aps: Generating captions for audios in the wild,''
\newblock in {\em Proceedings of the 2019 Conference of the North {A}merican
  Chapter of the Association for Computational Linguistics}. pp. 119--132, ACL.

\bibitem{Mei2021AnEB}
Xinhao Mei, Qiushi Huang, Xubo Liu, Gengyun Chen, Jingqian Wu, Yusong Wu,
  Jinzheng Zhao, Shengchen Li, Tom Ko, H~Lilian Tang, Xingkun Shao, MarkD~.
  Plumbley, and Wenwu Wang,
\newblock ``An encoder-decoder based audio captioning system with transfer and
  reinforcement learning,''
\newblock {\em ArXiv}, vol. abs/2108.02752, 2021.

\end{thebibliography}
\vspace{-5mm}

\end{document}